\documentclass{article}

     \PassOptionsToPackage{numbers, compress}{natbib}

\usepackage[preprint]{neurips_2022}

\usepackage[utf8]{inputenc} %
\usepackage[T1]{fontenc}    %
\usepackage[pagebackref=true,breaklinks=true,colorlinks,bookmarks=false,citecolor=ForestGreen]{hyperref}     %
\usepackage{url}            %
\usepackage{booktabs}       %
\usepackage{amsfonts}       %
\usepackage{nicefrac}       %
\usepackage{microtype}      %
\usepackage[dvipsnames]{xcolor}         %
\usepackage[pdftex]{graphicx}
\usepackage{amsmath}
\usepackage{amssymb}
\usepackage{multirow}
\usepackage[caption=false,position=top]{subfig}
\usepackage{colortbl}
\usepackage{dsfont}
\usepackage{pifont}
\usepackage{wrapfig}
\usepackage{titlesec}
\usepackage{xspace}
\usepackage{cleveref}
\usepackage{enumitem}
\usepackage{comment}
\usepackage{makecell}
\makeatletter\renewcommand\paragraph{\@startsection{paragraph}{4}{\z@}
	{.2em \@plus1ex \@minus.2ex}{-.5em}{\normalfont\normalsize\bfseries}}\makeatother
\newcommand{\app}{\raise.17ex\hbox{$\scriptstyle\sim$}}
\makeatletter
\DeclareRobustCommand\onedot{\futurelet\@let@token\@onedot}
\def\@onedot{\ifx\@let@token.\else.\null\fi\xspace}

\def\eg{\emph{e.g}\onedot} 
\def\ie{\emph{i.e}\onedot} 
 
 \def\vs{\emph{vs}\onedot}

\makeatother

\definecolor{baselinecolor}{gray}{.9}

\newcommand{\inc}[1]{{\color{blue}{#1}}}
\newcommand{\dec}[1]{{\color{red}{#1}}}
\newcommand{\unc}[1]{{\color{gray}{#1}}}

\title{Sight Beyond Text: Multi-Modal Training Enhances LLMs in Truthfulness and Ethics}

\author{%
  Haoqin Tu*$^1$\quad
  Bingchen Zhao*$^2$\quad
  Chen Wei$^3$ \quad
  Cihang Xie$^4$ \vspace{.3em}\\ 
  \small *equal contribution \vspace{.5em} \\
  $^1$ University of Chinese Academy of Sciences ~~ $^2$ University of Edinburgh \\ $^3$ Johns Hopkins University ~~
    $^4$ UC Santa Cruz \vspace{.3em} \\
}

\begin{document}

\maketitle

\begin{abstract}
Multi-modal large language models (MLLMs) are trained based on large language models (LLM), with an enhanced capability to comprehend multi-modal inputs and generate textual responses. While they excel in multi-modal tasks, the pure NLP abilities of MLLMs are often underestimated and left untested.
In this study, we get out of the box and unveil an intriguing characteristic of MLLMs --- our preliminary results suggest that visual instruction tuning, a prevailing strategy for transitioning LLMs into MLLMs, unexpectedly and interestingly helps models attain both improved truthfulness and ethical alignment in the pure NLP context. 
For example, a visual-instruction-tuned LLaMA2 7B model surpasses the performance of the LLaMA2-chat 7B model, fine-tuned with over one million human annotations, on \texttt{TruthfulQA-mc} and \texttt{Ethics} benchmarks. 
Further analysis reveals that the improved alignment can be attributed to the superior instruction quality inherent to visual-text data.
In releasing our code at \url{github.com/UCSC-VLAA/Sight-Beyond-Text}, we aspire to foster further exploration into the intrinsic value of visual-text synergies and, in a broader scope, multi-modal interactions in alignment research.

\end{abstract}

\section{Introduction}

As large language models (LLMs) have already catalyzed significant transformations across various fields, their evolution to multi-modal capabilities, facilitating interactions with inputs from various domains in a human-like way, has garnered substantial interest.
While prior research on multimodal LLMs (MLLMs) largely focuses on their prowess in tasks such as visual reasoning and visual-grounded generation~\cite{liu2023visual,zhu2023minigpt}, there remains a limited investigation for reassessing their performance on pure NLP tasks.

In this paper, we are interested in probing how multi-modal training can affect the ``hallucination'' phenomenon observed in LLMs, where they generate misleading or factually incorrect content. 
Our focus is primarily on visual instruction tuning, a technique that turns LLMs into MLLMs by tuning with visual-text inputs. Our findings are compelling --- visual instruction tuning intriguingly presents a promising pathway to enhance the truthfulness and ethics of LLMs.
For example, a vanilla LLaMA2 7B model, post visual instruction tuning, is able to register impressive scores of 46.0\% on \texttt{TruthfulQA-mc} (+7.1\%) ~\cite{lin2022truthfulqa} and 65.4\% on \texttt{Ethics} (+19.6\%) ~\cite{hendrycks2020aligning}, depending on the specific tuning approach. 
It is noteworthy that, even without engineering prompts that explicitly prioritize ethical or truthful behaviors, the performance of our visually-instructed model already outperforms that of the LLaMA2-chat 7B variant, which is heavily tuned with over a million human annotations~\cite{touvron2023llama2}. 
Moreover, our analysis suggests that the primary factor in driving such stronger alignments is the quality of visual-text instructions, 
and our ablations show that the efficacy of different forms of visual instruction data is intricately tied to the specific alignment standard measured.

\begin{figure}[t]
    \centering
    \includegraphics[width=.85\linewidth]{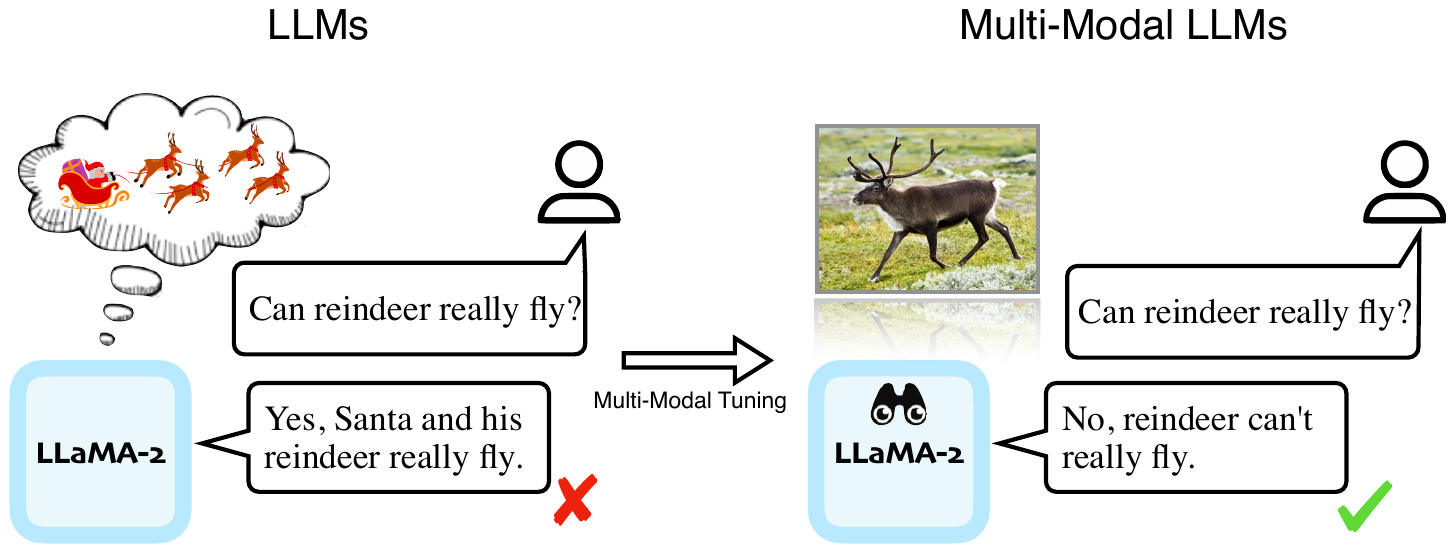}
    \vspace{-.7em}
    \caption{Visual instruction tuning substantially improves the truthfulness and ethics of LLMs. We observe that tuning LLMs with only 80k multi-modal data can yield stronger results on truthfulness and ethics than those with over one million human-annotated RLHF data.  
    Note that these LLMs employ images only during the visual instruction tuning and are tested without images for NLP tasks.}
    \vspace{-.5em}
    \label{fig:intro}
\end{figure}

In summary, our insights accentuate the promise of visual instruction tuning in 
fostering ethical and truthful alignment in LLMs. 
We present our findings as a cornerstone, inspiring the research community to explore the potential of visual instruction tuning, and, more broadly, multi-modal interactions in rectifying AI alignment conundrums.

\section{Tuning LLMs with Multi-Modal Data}
This section introduces our strategies to tune LLMs using multi-modal datasets. 
A standard MLLM typically contains three key components: 1) a vision encoder tasked with encoding visual inputs, 2) a vision-language connector that translates visual tokens into the linguistic space, and 3) an LLM for decoding the transcribed visual information. 
We strictly adhere to the setups in LLaVA~\cite{liu2023visual} for fine-tuning LLMs on visual instruction tuning data.

\paragraph{Model architecture.}
We incorporate the pre-trained visual branch of CLIP ViT-L/14~\cite{radford2021learning} as our vision encoder. Additionally, a trainable linear layer is employed to project visual tokens into the language embedding space. 
Regarding the choice of LLM, we take the widely recognized open-sourced LLaMA models~\cite{touvron2023llama1,touvron2023llama2,openlm2023openllama} for this study. 
Specifically, our investigation focuses on the following six models, containing three latest LLMs and their corresponding instruction-tuned variants:
\begin{itemize}
    \item Pre-trained LLM: OpenLLaMA-3B~\cite{openlm2023openllama}, LLaMA-7B~\cite{touvron2023llama1}, LLaMA2-7B~\cite{touvron2023llama2}.
    \item Instruction-tuned LLM: OpenAlpaca-3B~\cite{openalpaca}, Vicuna-7B~\cite{zheng2023judging}, LLaMA2-chat-7B~\cite{touvron2023llama2}.
\end{itemize}

As listed above, our study is centered on two model scales: 3B and 7B. 
While the 3B LLaMA model is sourced from the OpenLM project~\cite{openlm2023openllama}, the 7B LLaMA models are directly released by Meta~\cite{touvron2023llama1,touvron2023llama2}; additionally, our investigation extends to the instruction-tuned variants of these base LLMs. 
Concretely, OpenAlpaca-3B is fine-tuned on the Alpaca data~\cite{alpaca} using OpenLLaMA-3B as its backbone; 
Vicuna-7B is the v1.1 model from FastChat~\cite{zheng2023judging}, which is crafted upon LLaMA-7B and employs 125K conversational data from ShareGPT~\cite{sharegpt} during tuning; 
LLaMA2-chat-7B is well-engineered for human alignment, undergoing its training on publicly available instruction datasets and one million human-annotated examples using RLHF techniques. 
Note that we test 7B LLM variants by default, and indicate 3B models by the suffix ``-3B''.

\paragraph{Training procedure.}
The MLLM training unfolds in two stages. First, we exclusively tune the weight of the vision-language connector, with both the visual encoder and the LLM remaining frozen. In the second phase, we fine-tune the weights of both the connector and the LLM. Data-wise, we adhere to the protocols set by LLaVA~\cite{liu2023visual}:
the connector is initially trained using 595k image-text pairings filtered from CC3M~\cite{changpinyo2021conceptual}; the subsequent stage utilizes 80k instructions-following data from LLaVA~\cite{liu2023visual}, containing image-grounded conversation, image descriptions, and image-based complex reasoning tasks. 
Note that in the second stage, we probe the effects of both full fine-tuning and LoRA fine-tuning~\cite{hu2021lora}.

\section{Evaluations}

\subsection{Truthfulness and Ethics of MLLMs}
\label{subsec:alignment}
We report the evaluation results on the \texttt{TruthfulQA} and \texttt{Ethics} benchmarks, designed for measuring LLMs' truthfulness and ethical alignment. 
During this evaluation, we utilize the weights exclusively from the visual-instruction-tuned LLMs, intentionally omitting the visual encoders and vision-language connectors introduced during the fine-tuning process. 
The results are presented in~\cref{tab:eval_truth_ethics}.

\begin{table}[t]
\small
\centering
\begin{tabular}{l|c|ccc}
\toprule
Models           & \texttt{Ethics}         & \texttt{TruthfulQA-gen} & \texttt{TruthfulQA-mc1} & \texttt{TruthfulQA-mc2}             \\ \midrule
LLaMA           & 50.4\%                         & 27.5\%                 & 22.0\%                 &  34.1\%                  \\
MM-ft           & 59.1\% (\inc{+8.7\%})          & 29.4\% (\inc{+1.8\%})  & 23.6\% (\inc{+1.6\%})  &  35.8\% (\inc{+1.7\%})  \\
\midrule 
Vicuna            & 66.6\%                       & 45.4\%                 & 32.0\%                 &  47.0\%                  \\
MM-ft             & 60.4\% (\dec{-6.2\%})        & 29.3\% (\dec{-16.1\%}) & 23.8\% (\dec{-8.1\%})  &  35.8\% (\dec{-11.2\%}) \\
\midrule \midrule
LLaMA-3B          & 45.6\%                       & 25.3\%                 & 21.3\%                 &  34.6\%                  \\

MM-ft             & 58.1\% (\inc{+12.5\%})       & 26.4\% (\inc{+1.1\%})  & 21.4\% (\unc{+0.1\%})  &  32.9\% (\dec{-1.7\%})  \\

MM-lora           & 45.7\% (\unc{+0.1\%})       & 25.2\% (\unc{-0.1\%})  & 23.0\% (\inc{+1.7\%})  &  35.6\% (\inc{+1.0\%})  \\
\midrule

Alpaca-3B         & 44.0\%                       & 28.6\%                 & 22.4\%                 &  34.2\%                  \\
MM-ft             & 46.8\% (\inc{+2.8\%})        & 28.2\% (\unc{-0.4\%})  & 23.1\% (\inc{+0.7\%})  &  34.2\% (\unc{+0.0\%})  \\

MM-lora           & 44.0\% (\unc{+0.0\%})        & 28.6\% (\unc{+0.0\%})        & 24.6\% (\inc{+2.2\%})  &  38.0\% (\inc{+3.8\%})  \\
\midrule \midrule

LLaMA2            & 45.8\%                       & 32.3\%                 & 25.2\%                 &  38.9\%                  \\
MM-ft             & 65.4\% (\inc{+19.6\%})       & 31.5\% (\dec{-0.9\%})  & 27.8\% (\inc{+2.6\%})  &  40.2\% (\inc{+1.3\%})  \\

MM-lora           & 46.1\% (\unc{+0.3\%})        & 37.9\% (\inc{+5.6\%})  & 32.1\% (\inc{+6.9\%})  &  46.0\% (\inc{+7.1\%})  \\
\midrule
LLaMA2-chat       & 58.5\%        & 43.3\%          & 29.5\%  &  44.6\%                  \\

MM-ft             & 65.2\% (\inc{+6.7\%})        & 35.5\% (\dec{-7.8\%})  & 27.7\% (\dec{-1.8\%})  &  41.0\% (\dec{-3.6\%})  \\

MM-lora           & 58.6\% (\unc{+0.1\%})        & 44.6\% (\inc{+1.2\%})  & 29.4\% (\unc{-0.1\%})  &  44.6\% (\unc{+0.0\%})  \\

\bottomrule
\end{tabular}
\vspace{2.5pt}
\caption{Comparison on the original LLMs and the multi-modal fine-tuned ones on \texttt{Ethics}~\cite{hendrycks2020aligning} and \texttt{TruthfulQA}~\cite{lin2022truthfulqa}. `-ft' represents full parameter fine-tuning and `-lora' indicates LoRA tuning. We report Rouge-L accuracy for \texttt{TruthfullQA-gen} and accuracy for the rest.}
\label{tab:eval_truth_ethics}
\end{table}

\paragraph{Visual instruction tuning improves truthfulness and ethics.}
Our observations suggest that, rather unexpectedly, visual instruction tuning tends to enhance the truthfulness of LLMs. A compelling observation emerges when comparing LLaMA2 variants: visual-instruction-tuned models, especially LLaMA2 with MM-lora, surpass the LLaMA2-chat model in performance metrics on both \texttt{TruthfulQA-mc1} (32.1\% \vs 29.5\%) and \texttt{TruthfulQA-mc2} (46.0\% \vs 44.6\%).

From~\cref{tab:eval_truth_ethics}, we also observe visual instruction tuning leads to substantial improvements on the \texttt{Ethics} task. Echoing the trend in the \texttt{TruthfulQA} evaluations, visual-instruction-tuned models, specifically MM-ft versions of both LLaMA2 and LLaMA-3B, consistently outpace their instruction-tuned counterparts, such as LLaMA2-chat and Alpaca-3B. For example, the performance enhancements observed for LLaMA2 and LLaMA-3B on the \texttt{Ethics} task amounted to increments of 19.6\% and 12.5\% respectively, outperforming LLaMA2-chat and Alpaca-3B by margins of 6.9\% and 11.3\%.

It should be noted that the employed visual instruction tuning data is the 80k dataset derived from LLaVA~\cite{liu2023visual}, which does not contain special designs for aligning models to human preferences. 
Remarkably, despite this, visual instruction tuning is able to yield empirical advantages that surpass those from RLHF, which heavily utilizes a substantial corpus of human-annotated data dedicated to LLM alignment.
This observation strongly attests to the potential that visual instruction tuning holds in addressing AI alignment challenges. 
However, it is not a silver bullet --- our experiments also show that visual instruction tuning is limited at enhancing the alignment of models previously fine-tuned via instruction tuning (\eg, models like Vicuna, LLaMA2-chat), indicating a variability in its efficacy.

\begin{figure}[t]
    \centering
    \includegraphics[width=.97\linewidth]{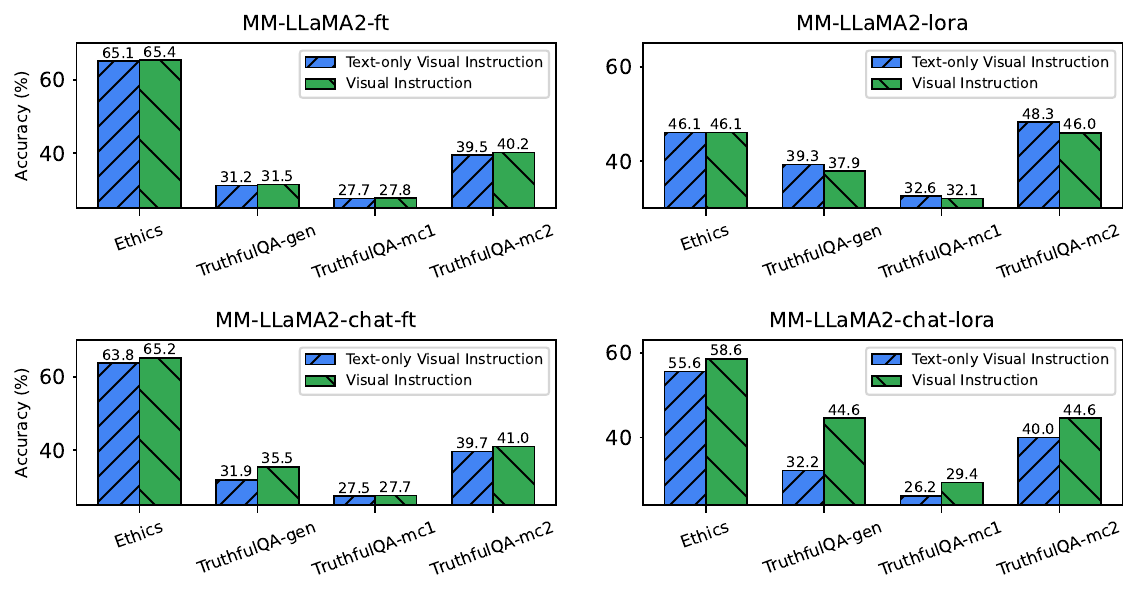}
    \vspace{-1.1em}
    \caption{Performance of visual-instruction-tuned LLaMA2 models and text instruction tuned ones on \texttt{Ethics} and \texttt{TruthfulQA} benchmarks. The text-only visual instruction data is taken directly from LLaVA, but without the paired images.}
    \label{fig:text_image_ablation2}
\end{figure}

\begin{figure}[t]
    \centering
    \includegraphics[width=.97\linewidth]{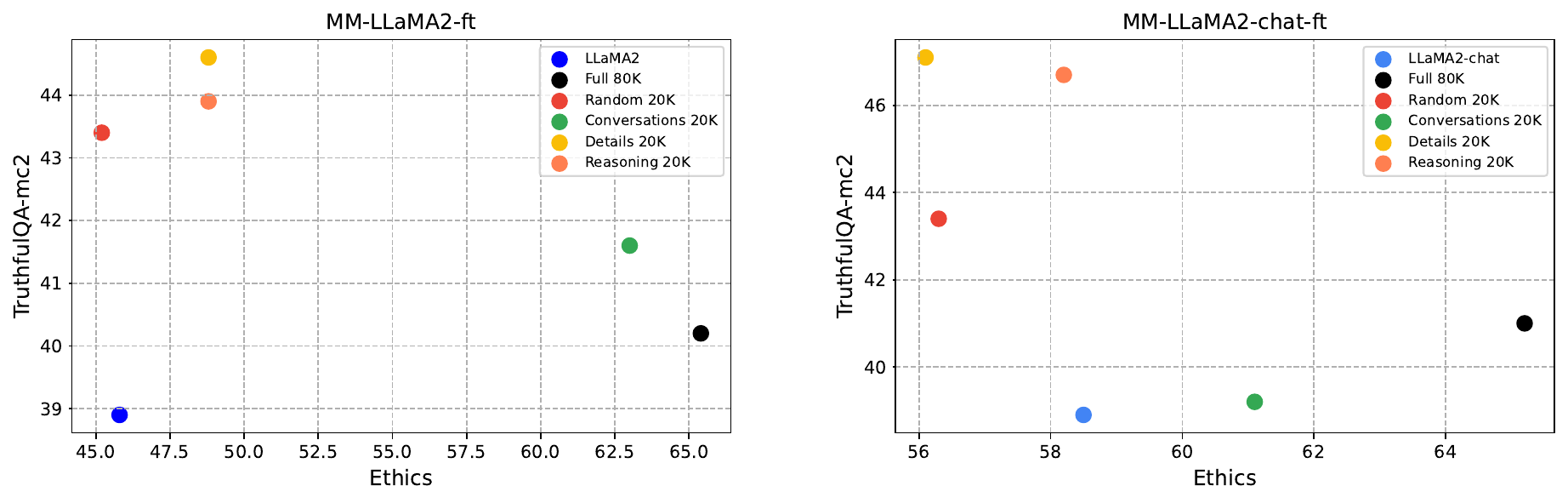}
    \vspace{-1em}
    \caption{Results of different data components on \texttt{Ethics} and \texttt{TruthfulQA} of visual-instruction-tuned LLMs. We utilize 20K of different forms of data (Conversation, Details, Reasoning), and additionally sample 20K data out of the original 80K training instances (Random 20K) for comparison.}
    \label{fig:data_component}
\end{figure}

\paragraph{Effects of modalities in visual instruction-tuning data on LLM alignment.}
Next, we seek to understand how different modalities in the visual instruction data contribute to the alignment of LLMs. Specifically, we design a set of ablations where we only utilize the text part of the visual instruction tuning data to tune the LLMs, and draw a comparison with the models tuned with both the visual inputs and the corresponding texts.

As shown in~\cref{fig:text_image_ablation2}, we observe that models with text-only visual instruction tuning can largely attain comparable alignment performance with the vanilla visual instruction tuning baseline where both images and texts are used.
While additionally including visual inputs yields seemingly ``modest'' alignment improvements, we stress that these gains are consistent across different LLMs, tuning methods, and alignment tasks. For example, this can be verified across three model variants, resulting in an average accuracy improvement of 2.5\% across three sub-tasks presented in~\cref{fig:text_image_ablation2}.

This observation leads to our hypothesis that there exists a promising avenue in leveraging visual data to construct enhanced instruction-tuning datasets.
Although textual information plays a significant role in alignment, it is crucial to recognize that this text is inherently grounded in its corresponding real-world visual content; therefore, utilizing such paired information is integral to ensuring strong alignment in LLMs. These findings underscore the multifaceted benefits of visual data: it not only enhances alignment quality but also contributes significantly to the creation of more accurate instruction-tuning datasets.

\paragraph{Types of visual instruction data matters.}
We further extend our investigation to understand how varying types of visual instruction-tuning data affect LLM alignment.
Specifically, we utilize data from LLaVA~\cite{liu2023visual}, which categorizes visual instruction tuning data into three groups: \texttt{Conversation}, \texttt{Details}, and \texttt{Reasoning}. Each group comprises 20k data points, sampled from the original training splits. For a fair comparison, we also take a uniform sample of 20k from the full 80k visual instructions to form the baseline group.
We tune LLaMA2 and LLaMA2-chat with each data group (of 20k data points) separately, and report the results in \cref{fig:data_component}.

Our analysis reveals that, in general, conversational data has a greater impact on improving LLMs' performance on the \texttt{Ethics} task, resulting in an improvement of \app15\%  on MM-LLaMA2-ft and \app3\% on MM-LLaMA2-chat-ft. Conversely, reasoning and details data tend to be more effective in improving performance on the \texttt{TruthfulQA} benchmark, yielding gains of more than 2\% and 6\% on these two models.
This suggests that a targeted approach, leveraging the unique strengths of each data type, can facilitate more nuanced and effective instruction tuning for LLM alignment.

\subsection{Standard NLP Abilities}

\begin{table}[t]
\setlength\tabcolsep{3.5pt}
\small
\centering
\begin{tabular}{l|ccccc}
\midrule
Models                    & \texttt{MMLU} (Acc.)          & \texttt{GSM8K} (Acc.)         & \texttt{MathQA} (Acc.)        & \texttt{sQuAD} (F1.)          & \texttt{BoolQ} (Acc.)          \\ \midrule
LLaMA                     & 36.8\%         & 8.0\%          & 27.7\%         & 19.5\%          & 75.1\%          \\
MM-ft               & 27.7\% (\dec{-9.0\%}) & 0.9\% (\dec{-7.1\%})  & 28.5\% (\inc{+0.8\%}) & 9.1\% (\dec{-10.4\%})  & 47.5\% (\dec{-27.6\%}) \\
\midrule
Vicuna                    & 47.2\%         & 10.0\%         & 29.0\%         & 19.3\%          & 78.1\%          \\
MM-ft              & 44.0\% (\dec{-3.2\%}) & 5.4\% (\dec{-4.6\%})  & 29.4\% (\unc{+0.4\%}) & 10.1\% (\dec{-9.2\%})  & 52.5\% (\dec{-25.6\%}) \\
\midrule
LLaMA-3B                  & 26.7\%         & 2.4\%          & 26.4\%         & 20.7\%          & 65.6\%          \\
MM-ft         & 26.5\% (\unc{-0.2\%}) & 1.7\% (\dec{-0.6\%})  & 25.8\% (\dec{-0.6\%}) & 8.6\% (\dec{-12.1\%})  & 53.6\% (\dec{-12.0\%}) \\
\cellcolor[rgb]{0.929,0.929,0.929}{MM-lora}       & \cellcolor[rgb]{0.929,0.929,0.929}{26.8\% (\unc{+0.1\%})} & \cellcolor[rgb]{0.929,0.929,0.929}{3.1\% (\inc{+0.7\%})} & \cellcolor[rgb]{0.929,0.929,0.929}{26.3\% (\unc{-0.1\%})} & \cellcolor[rgb]{0.929,0.929,0.929}{18.8\% (\dec{-1.9\%})}  & \cellcolor[rgb]{0.929,0.929,0.929}{66.3\% (\inc{+0.7\%})}  \\
\midrule
Alpaca-3B                 & 24.9\%         & 0.1\%          & 24.6\%         & 28.2\%          & 71.1\%          \\
MM-ft        & 24.5\% (\unc{-0.4\%}) & 0.0\% (\unc{-0.1\%})  & 25.6\% (\inc{+1.0\%}) & 12.1\% (\dec{-16.1\%}) & 69.0\% (\dec{-2.1\%})  \\
\cellcolor[rgb]{0.929,0.929,0.929}{MM-lora}      & \cellcolor[rgb]{0.929,0.929,0.929}{24.3\% (\dec{-0.6\%})} & \cellcolor[rgb]{0.929,0.929,0.929}{0.1\% (\unc{+0.0\%})} & \cellcolor[rgb]{0.929,0.929,0.929}{25.4\% (\inc{+0.8\%})} & \cellcolor[rgb]{0.929,0.929,0.929}{24.2\% (\dec{-4.1\%})}  & \cellcolor[rgb]{0.929,0.929,0.929}{71.1\% (\unc{+0.0\%})}  \\
\midrule
LLaMA2                    & 45.9\%         & 13.7\%         & 30.1\%         & 26.3\%          & 77.7\%          \\
MM-ft        & 39.4\% (\dec{-6.5\%}) & 5.5\% (\dec{-8.2\%})  & 29.6\% (\unc{-0.5\%}) & 8.5\% (\dec{-17.8\%})  & 56.3\% (\dec{-21.4\%}) \\

\cellcolor[rgb]{0.929,0.929,0.929}{MM-lora}      & \cellcolor[rgb]{0.929,0.929,0.929}{46.6\% (\inc{+0.7\%})} & \cellcolor[rgb]{0.929,0.929,0.929}{15.0\% (\inc{+1.3\%})} & \cellcolor[rgb]{0.929,0.929,0.929}{30.4\% (\unc{+0.3\%})} & \cellcolor[rgb]{0.929,0.929,0.929}{20.1\% (\dec{-6.2\%})}  & \cellcolor[rgb]{0.929,0.929,0.929}{77.6\% (\unc{-0.1\%})}  \\
\midrule

LLaMA2-chat               & 45.8\%         & 18.2\%         & 31.1\%         & 20.1\%          & 80.7\%          \\
MM-ft   & 45.2\% (\dec{-0.6\%}) & 6.2\% (\dec{-12.0\%}) & 30.0\% (\dec{-1.1\%}) & 10.2\% (\dec{-9.3\%})  & 67.0\% (\dec{-13.7\%}) \\

\cellcolor[rgb]{0.929,0.929,0.929}{MM-lora} & \cellcolor[rgb]{0.929,0.929,0.929}{45.9\% (\unc{+0.2\%})} & \cellcolor[rgb]{0.929,0.929,0.929}{17.1\% (\dec{-1.1\%})}  & \cellcolor[rgb]{0.929,0.929,0.929}{30.8\% (\unc{-0.3\%})} & \cellcolor[rgb]{0.929,0.929,0.929}{25.5\% (\inc{+5.4\%})}  & \cellcolor[rgb]{0.929,0.929,0.929}{81.5\% (\inc{+0.8\%})}  \\ \midrule
\end{tabular}
\caption{Performances of both the vanilla LLMs and visual-instruction-tuned LLMs on five NLP capabilities benchmarks.}
\label{tab:nlp}
\vspace{-1.8em}
\end{table}

Given these LLMs are further fine-tuned with multi-modal data, it might be intuitively expected that their standard NLP capabilities could degrade. Such a phenomenon is commonly referred to as catastrophic forgetting~\cite{kirkpatrick2017overcoming} or in the AI alignment community --- the alignment tax \cite{alignment_tax1,alignment_tax2}.

Interestingly, contrary to these assumptions, our results presented in~\cref{tab:nlp} show that MM-lora (marked in the gray background) results in only an average 0.17\% performance decrease across five NLP capability benchmarks and four models, after applying visual instruction tuning. More notably, in certain instances, MM-lora even modestly improves performance on these benchmarks.

In conjunction with the insights from Section~\ref{subsec:alignment}, these observations altogether highlight the ability of visual-instruction-tuned LLMs in both maintaining the strong capability on standard NLP benchmarks and aligning better with human values, not to mention the additional capability of recognizing visual inputs. Such findings pave new avenues for both academic exploration and practical implementations within multi-modal domains. We believe these insights should catalyze further investigations into the tuning of LLMs with multi-modal interactions.

\subsection{Analysis on Multi-Modal Benchmarks}
\label{sec:mm}

\begin{table}[htb]
\setlength\tabcolsep{3.5pt}
\small
\centering
\begin{tabular}{l|cccccc}
\midrule
Models          & \texttt{MME CS}          & \texttt{MME PS}          & \texttt{VQAv2}       & \texttt{MSCOCO}           & \texttt{Flickr30k}  & \texttt{POPE} R / A / P    \\ \midrule
MM-LLaMA-ft             & 199.3          & 510.5          & 15.2             & 59.2             & 27.1     & 65.7 / 57.8 / 59.9     \\
MM-Vicuna-ft        & \textbf{270.7}          & 625.2          & 15.4             & 57.5             & 24.6    & \textbf{76.5 / 66.5 / 73.8}      \\
MM-LLaMA2-ft        & {237.1} & 661.3          & \textbf{16.1}  & \textbf{65.1}  & \textbf{31.6}  & 65.0 / 55.4 / 56.3 \\
MM-LLaMA2-lora      & 200.0          & 395.0          & 14.9             & 52.0             & 26.2      & 50.8 / 50.4 / 50.6    \\
MM-LLaMA2-chat-ft      & 234.6          & \textbf{805.4} & 15.3             & 57.4             & 26.7    & 69.8 / 57.9 / 60.3      \\
MM-LLaMA2-chat-lora & 228.6          & 709.8          & 13.8             & 43.4             & 23.0   & 65.9 / 56.8 / 59.2 \\
\bottomrule
\end{tabular}
\caption{Performances of our MLLM family on five widely employed multi-modal benchmarks. On \texttt{POPE}, we test all models on three sub-tasks: Random (R), Adversarial (A), and Popular (P).}
\label{tab:general_mm_bench}
\end{table}

\begin{table}[ht]
\setlength\tabcolsep{3.5pt}
\small
\centering
\begin{tabular}{l|cc}
\midrule
Models                & \texttt{MSCOCO} (CIDEr)        & \texttt{MSCOCO-C} (CIDEr)     \\ \midrule
MM-LLaMA-ft              & 59.2          & 48.6 (\dec{-17.9\%})                  \\
MM-Vicuna-ft             & 57.5          & 46.0 (\dec{-20.0\%})                    \\
MM-LLaMA2-ft          & \textbf{65.1}          & \textbf{54.6} (\dec{-16.1\%})  \\
MM-LLaMA2-lora        & 52.0          & 43.2 (\dec{-16.9\%})            \\
MM-LLaMA2-chat-ft     & 57.4          & 47.5 (\dec{-17.2\%})             \\
MM-LLaMA2-chat-lora   & 43.4          & 33.8 (\dec{-22.1\%})              \\ \midrule       
\end{tabular}
\caption{Performances of the MLLM family on \texttt{MSCOCO}~\cite{lin2014microsoft} with corrupted visual inputs.}
\label{tab:corrupted_mm_bench}
\vspace{-1em}
\end{table}

In this section, we test the visual-instruction tuned models on recent multi-modal evaluation benchmarks, where five multi-modal benchmarks are deployed: \texttt{MME}~\cite{fu2023mme} consists of two evaluation aspects, \ie, cognition (\texttt{CS}) and perception (\texttt{PS}) with total 14 VQA tasks;\footnote{We exclude \texttt{landmark} and \texttt{artwork} tasks to accelerate the evaluation process.} \texttt{VQAv2}~\cite{antol2015vqa}, \texttt{MSCOCO}~\cite{lin2014microsoft} and \texttt{Flickr30k}~\cite{young2014image} captioning tasks are commonly used benchmarks in the field of VQA and captioning. The former two benchmarks are based on \texttt{MSCOCO}-2017 dataset~\cite{lin2014microsoft}. For the latter two captioning tasks, we report the zero-shot CIDEr~\cite{vedantam2015cider} scores (with three text-only QA examples) on the test set from the Karpathy split~\cite{karpathy2015deep}. 
Additionally, We also make use of the image corruptions proposed in \texttt{ImageNet-C}~\cite{hendrycks2019robustness} to measure the performance of the MLLMs on corrupted images for \texttt{MSCOCO} tasks (denoted as \texttt{MSCOCO-C}).\footnote{For corrupted images, we report the average results of tested models on four noises (gaussian noise, defocus blur, contrast, brightness) across three severity levels (1, 3, 5)}. 
\texttt{POPE}~\cite{li2023evaluating} is used to evaluate the level of object hallucinations in MLLMs, which consists of three versions of balanced yes/no VQA tasks considering objects in the given image.

\paragraph{Potential inconsistency in current multi-modal benchmarks.}
In~\cref{tab:general_mm_bench}, MLLMs incorporating text-aligned LLMs have demonstrated superior performance in comprehensive and challenging tasks such as \texttt{MME} and \texttt{POPE}. Specifically, MM-Vicuna-ft and MM-LLaMA-chat-ft outperform their corresponding vanilla MLLM counterparts by an average of 164.9 on \texttt{MME} and 7.5\% on \texttt{POPE}. However, despite the incorporation of text-aligned LLMs, MLLMs exhibit unexpected shortcomings in comparison to models leveraging vanilla LLMs when evaluated on three traditional vision-text tasks (\eg, an average 4.2 CIDEr score drop on two captioning tasks). The inconsistent performance comparison across these five benchmarks highlights the imperative for improving evaluation techniques within multi-modal benchmarks.

\paragraph{Need for studying multi-modal alignments.}
Despite the effectiveness of text-aligned models like Vicuna and LLaMA2-chat, their MLLM variants exhibit poor performance on corrupted images as shown in~\cref{tab:corrupted_mm_bench}. These models not only lag behind MLLMs without instruction-tuned LLMs, but also demonstrate performance drops of over 17\% when evaluated on corrupted images compared to clean ones, which are higher than drops observed for MM-LLaMA-ft and MM-LLaMA2-ft.
This observation indicates that though visual instruction tuning improves the truthfulness and ethics of LLMs in the language domain, these MLLMs still face their unique challenges in the multi-modal domain.

\section{Related Work}
\paragraph{Alignments.}
The alignment of AI systems to human values is an important topic for today's advanced AI systems, from testing model robustness to out-of-distribution shifts~\cite{hendrycks2019robustness,hendrycks2021many,zhao2022ood} to adversarial attacks~\cite{hendrycks2021nae,eykholt2018robust,xie2020adversarial}, many works have been proposed.
The recent development of LLMs has revolutionized natural language processing and has been widely adopted in various applications.
Thus, concerns regarding the honesty and truthfulness of these models have also emerged, prompting alignment researchers to investigate the ethical implications and potential risks associated with their deployment. 
\texttt{TruthfulQA}~\cite{lin2022truthfulqa} is proposed to measure how LLMs imitate human misconceptions. 
And \texttt{Ethics}~\cite{hendrycks2020aligning} is used to assess a language model’s knowledge of basic concepts of
morality. 
Given the popularity of the use of large language models, adversarial attacks on LLMs have also been explored~\cite{zou2023universal}.
In this work, we present our findings on how visual instruction tuning can help the LLMs align with human values, our results show impressive performance boost on these datasets without explicit prompting such behaviors.

\paragraph{Multi-Modal and Large Language Models.}
Multi-modality has long been a hot topic, CLIP~\cite{radford2021learning} proposes to align representations of both images and text, and later works proposed more techniques for this aim~\cite{yu2022coca,mu2022slip,zhao2023vision}.
In light of the rapid evolvement of large language models (LLMs), recent studies about multi-modal systems have turned their focuses from incorporating fine-grained multi-modal data~\cite{liang2021maria,tu2023resee} to integrating powerful LLMs with few-shot capability. 
More recently, some instruction-tuned MLLMs have emerged, showing excellent generalization ability in unseen VL tasks~\cite{zhu2023minigpt,liu2023visual,ye2023mplug,li2023blip,Dai2023InstructBLIPTG}.
For example, MiniGPT4~\cite{zhu2023minigpt} is built upon QFormer~\cite{li2023blip} and Vicuna~\cite{zheng2023judging} and only activates the linear layer connecting the vision encoder and LLM. 
LLaVA~\cite{liu2023visual} projects the output of a vision encoder to word tokens and trains both the VL connector and the LLM on synthetic data. 
mPLUG-owl~\cite{ye2023mplug} tunes LLaMA with a query-based VL connector using both text-only and vision-language instruction data. 
InstructBLIP~\cite{Dai2023InstructBLIPTG} uses BLIP2~\cite{li2023blip} as the backbone but is additionally instruction-tuned on a collection of VL datasets.
In our work, we demonstrate a new perspective on these MLLMs -- tuning LLMs with multi-modal data greatly helps align them with human values.

\section{Conclusion, Discussion, and Future Work}

In this study, we offer preliminary findings that underscore the potential of enhancing the truthfulness and ethical alignment of LLMs through visual instruction tuning. Remarkably, even without prompts tailored for truthfulness or ethical behaviors, our approach to tuning LLM weights using visual instruction datasets yielded significant improvements in both the \texttt{TruthfulQA} and \texttt{Ethics} benchmarks. Notably, such improvements are even stronger than that of RLHF, which tunes LLMs with a huge corpus of human-aligned data points. 
The follow-up analysis demonstrates the importance of instruction data quality for improving aligned values in MLLMs, as well as specific types of data models employed for applying to different alignment tasks.

In light of our findings, we advocate for future research endeavors to focus on devising innovative methodologies for crafting visual instruction tuning data that can more effectively align LLMs. Exploring novel MLLM architectures could also be a fruitful avenue. We hope fostering LLM interactions with real-world environments may emerge as a pivotal strategy for achieving superior model alignment.

\section*{Acknowledge}
This work is partially supported by a gift from Open Philanthropy. We thank Center for AI Safety for supporting our computing needs.

{
\small
\bibliographystyle{plain}
\bibliography{example}
}

\end{document}